\documentclass{article}

\usepackage[numbers,compress]{natbib}

\usepackage[utf8]{inputenc} 
\usepackage[T1]{fontenc}    
\usepackage{lmodern}
\usepackage{hyperref}       
\usepackage{url}            
\usepackage{booktabs}       
\usepackage{amsfonts}       
\usepackage{nicefrac}       
\usepackage{microtype}      
\usepackage{xcolor}         
\usepackage{graphicx}       
\usepackage{authblk}        
\usepackage{amsmath}        

\usepackage{caption}
\usepackage[roman]{parnotes}
\usepackage{tabularx}

\usepackage{caption}        
\usepackage{subcaption}     

\usepackage{color,soul}     

\title{Expanding memory in recurrent spiking networks}

\author[$\dagger\ddag$]{Ismael Balafrej}
\author[$\ddag\star\dagger\dagger$]{Fabien Alibart}
\author[$\dagger\ddag$]{Jean Rouat}

\affil[$\dagger$]{
    NECOTIS Research Lab,
    Université de Sherbrooke, Canada
}

\affil[$\ddag$]{
    Institut interdisciplinaire d'innovation technologique,
    Université de Sherbrooke, Canada
}

\affil[$\star$]{
    Laboratoire Nanotechnologies et Nanosystèmes,
    Université de Sherbrooke, Canada
}

\affil[$\dagger\dagger$]{
    Institut d'électronique, de microélectronique et de nanotechnologie,
    Université de Lille, France
}

 \date{October 2022}

\newcommand{\he}{\ensuremath{\mathcal{H}}}
\newcommand{\re}{\ensuremath{\mathcal{R}}}

\newcommand{\thresh}{\ensuremath{\text{th}}}

\begin{document}

\maketitle

\begin{abstract}
    Recurrent spiking neural networks (RSNNs) are notoriously difficult to train because of the vanishing gradient problem that is enhanced by the binary nature of the spikes. In this paper, we review the ability of the current state-of-the-art RSNNs to solve long-term memory tasks, and show that they have strong constraints both in performance, and for their implementation on hardware analog neuromorphic processors. We present a novel spiking neural network that circumvents these limitations. Our biologically inspired neural network uses synaptic delays, branching factor regularization and a novel surrogate derivative for the spiking function. The proposed network proves to be more successful in using the recurrent connections on memory tasks. 
\end{abstract}

\section{Introduction}

Undoubtedly, one of the largest unexplained exploits of the human brain is its ability to memorize useful patterns over very long timescales. Neural networks based on the transformers architecture \citep{Vaswani2017-dw} have been becoming increasingly good at processing temporal tasks and other sequence-like data. However, these models rely on self-attention layers that possess a quadratic computational complexity of the length of a buffered sequence. This comes at a huge energy cost and highly limits the processing capabilities of very long sequences. Recurrent neural networks (RNN) process sequences one step at a time, with an internal hidden state to represent past information. They have a constant memory requirement, and they scale linearly when trained with backpropagation through time (BPTT), which is the most common way to train RNNs~\citep{Schmidhuber2015-ok}.

Spiking neurons have emerged as an energy-efficient alternative to artificial neurons because of their event-based transmission. A spiking neuron only emits a binary output, or spike, when the accumulated temporal information reaches a threshold. The leaky-integrate-and-fire (LIF) neuron is the most commonly used model for spiking neurons, due to its simplicity and ease of hardware implementation~\citep{Han2022-gm}. Recurrent neural networks (RNN) are notoriously difficult to train because of the vanishing and exploding gradient problems~\citep{Bengio1994-lp}. The temporally sparse and binary network activity of spiking neural networks only adds to the training difficulty~\citep{Neftci2019-jq}. 

\begin{figure}[h]
    \centering
    \includegraphics[width=0.4\textwidth]{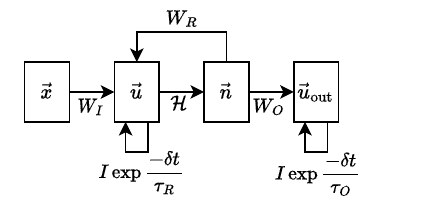}
    \caption{Schematic of a recurrent spiking neural network with leaky-integrate-and-fire (LIF) neurons. $W_R$, $\tau_O$ and $\tau_R$ are the parameters that impact the length of memory storage, both forward and backward. $I$ is the identity matrix. See text for details.}
    \label{fig:schematic_representation}
\end{figure}

\autoref{fig:schematic_representation} presents a schematic overview of a recurrent spiking neural network with LIF neurons. The memory of the network is held in two components, which are the membrane potential leakage of the recurrent and output layers, $\vec{u}$ and $\vec{u}_\text{out}$ respectively, and the recurrent connections $W_R$. The current solution to alleviate the binary aspect of spikes and ease the training during BPTT is to make the time constant $\tau_R$ of the LIF neuron's membrane potential arbitrary long, to match the length of the input sequence \citep{Frenkel2022-ij}. It is also possible to rely on another long decaying exponential as an additional state, as done in \citep{Bellec2020-ui} with an adaptive threshold that decays exponentially over long timescales. Alternatively, the time constant at the output layer, $\tau_O$, may also be increased to match the duration of the input samples~\citep{Rossbroich2022-xn}. In these schemes, the ability to solve a task is therefore tied to that time constant, as it enforces a flow of the gradient through $I\exp \frac{-\delta t}{\tau_R}$ (or $\tau_O$) with $I$ being the identity matrix.

One of the interesting aspects of the LIF neuron is that the exponential decay can be implemented with a single capacitor working in an analog regime. Analog computing is thought to be a keystone of the future of machine learning~\citep{Hinton2022-ql}. Memory access of digital systems comes at a huge energy and latency cost, often referred to as the von Neumann bottleneck. According to the current approach for RSNN learning, large leak time constants $\tau_R$, and therefore big capacitors, are required to have long-range memory. This requires the use of large and costly electronic circuits that can not subsequently be scaled down. Other type of technologies could reduce the scalability problem of capacitors, such as the resistivity drift of Phase Change Memories \citep{Demirag2021-vt}, but the research is still ongoing.

Liquid state machines \citep{Maass2002-yv} and other types of reservoirs with spiking neurons \citep{Balafrej2022-ym,Ivanov2021-ha} have been thought to be an alternative. In these networks, the recurrent weights are not trained using a supervised gradient-based algorithm, but rather sampled from a random distribution that attempts to maximize their theoretical memory, with optionally added biologically plausible plasticity rules in more recent works \citep{Jin2016-nt,Ivanov2021-ha,Balafrej2022-ym,Payvand2022-kw}. In many cases, they still fall short in terms of scalability and machine learning performance to their backpropagated spiking neural network counterparts. 

In this work, we empirically prove that prior surrogate gradient descent methods rely on large time constants not only for training, but also during inference as a memory for the network. This showcases the strong limitation of prior networks to learn sequences of longer timescales. We then propose DelayNet, a RSNN that implements a novel biologically inspired regularization factor and synaptic transmission delays to solve multisecond-long tasks without increasing the leakage time constant of the neurons. Our synaptic delays are more than an order of magnitude smaller, and they can be shared amongst neurons to facilitate future hardware implementations. DelayNet can solve long memory tasks over a period of time much longer than prior recurrent spiking neural networks trained with BPTT.

\section{Background}
The leaky integrate and fire (LIF) neuron model consists of a membrane potential $u(t)$, decaying exponentially by a factor of constant $\tau$:
\begin{equation}
    \frac{du(t)}{dt}=\frac{-u(t)}{\tau}+x(t)
    \label{eq:lif}
\end{equation}
where $x(t)$ is the sum of all the input currents and biases being fed to that neuron. When $u(t)$ reaches a threshold $u_\thresh$, the neuron emits a spike, which is assumed here to be binary. This behavior can be modeled using the Heaviside step function as an activation function for the neuron:
\[
    n(t) = \he(u(t)-u_\thresh)
\]
where $n(t)$ is the output of the neuron at time $t$. As the binary spike activation function is non-differentiable, it is common practice to use a surrogate gradient function during backpropagation. We refer to \citep{Neftci2019-cv} for a review of commonly used surrogate functions and \cite{Zenke2021-zf,Herranz-Celotti2022-gm} for empirical comparisons between them.

To train a network, \autoref{eq:lif} is first converted to its numerical solution with timestep $\delta t$:
\[
    u[t]=\exp\left(\frac{-\delta t}{\tau}\right)u[t-\delta t]+x[t]
\]

We use the simplest definition of a recurrent spiking neural network (RSNN). This RSNN is composed of 3 fully-connected layers: $W_i$ mapping the input to the latent space, $W_r$ self-connecting the latent space and $W_o$ connecting the latent space to the output. $\xi$ represents the error computed for a specific task out of the output layer. As common practice, the last layer is composed of non-spiking leaky integrator neurons~\citep{Zenke2021-zf,Bellec2020-ui} such that the analog value of the membrane potential better maps classification probabilities and values for regression tasks.

\subsection{Analysis of the computational graph of a RNN during BPTT}
\begin{figure}[h]
    \centering
    \includegraphics[width=0.7\textwidth]{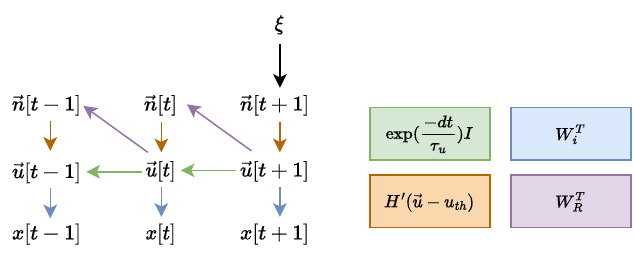}%
    \caption[Computational graph of a recurrent neural network unraveled for three timesteps]{Computational graph of a recurrent neural network unraveled for three timesteps during BPTT. Each arrow represents the local derivative given by the chain rule during BPTT, with each derivative presented at the right with the corresponding color. For the gradient to reach a previous state, e.g. $\vec{u}[t]$, it either has to go through a combination of the surrogate derivative $\he'$ in orange and the recurrent connections in violet (referred to as path 1), or through the derivative of the membrane potential leakage in green (referred to as path 2). $\xi$ represents the downstream error computed for a specific task with input $x[t]$.}%
    \label{fig:bptt_flow}
\end{figure}

During backpropagation through time (BPTT), the network is unraveled as if it was simply a very deep network with shared weights. \autoref{fig:bptt_flow} depicts the computational graph that is used during backpropagation through time for three timesteps. There are two distinct paths where the gradient can flow to train this network: (1) through the recurrent connections via the surrogate derivative $\he'$ - presented in orange and violet, or (2) through the membrane potential leakage - presented in green. Combining paths (1) and (2) can help with the vanishing gradient problem. Let us first look at each path individually.

\subsubsection{BPTT using the LIF neuron's leakage time constant}
\label{subsection:second_memory_lif_neuron}
\begin{figure}[h]
    \centering
    \includegraphics[width=0.5\textwidth]{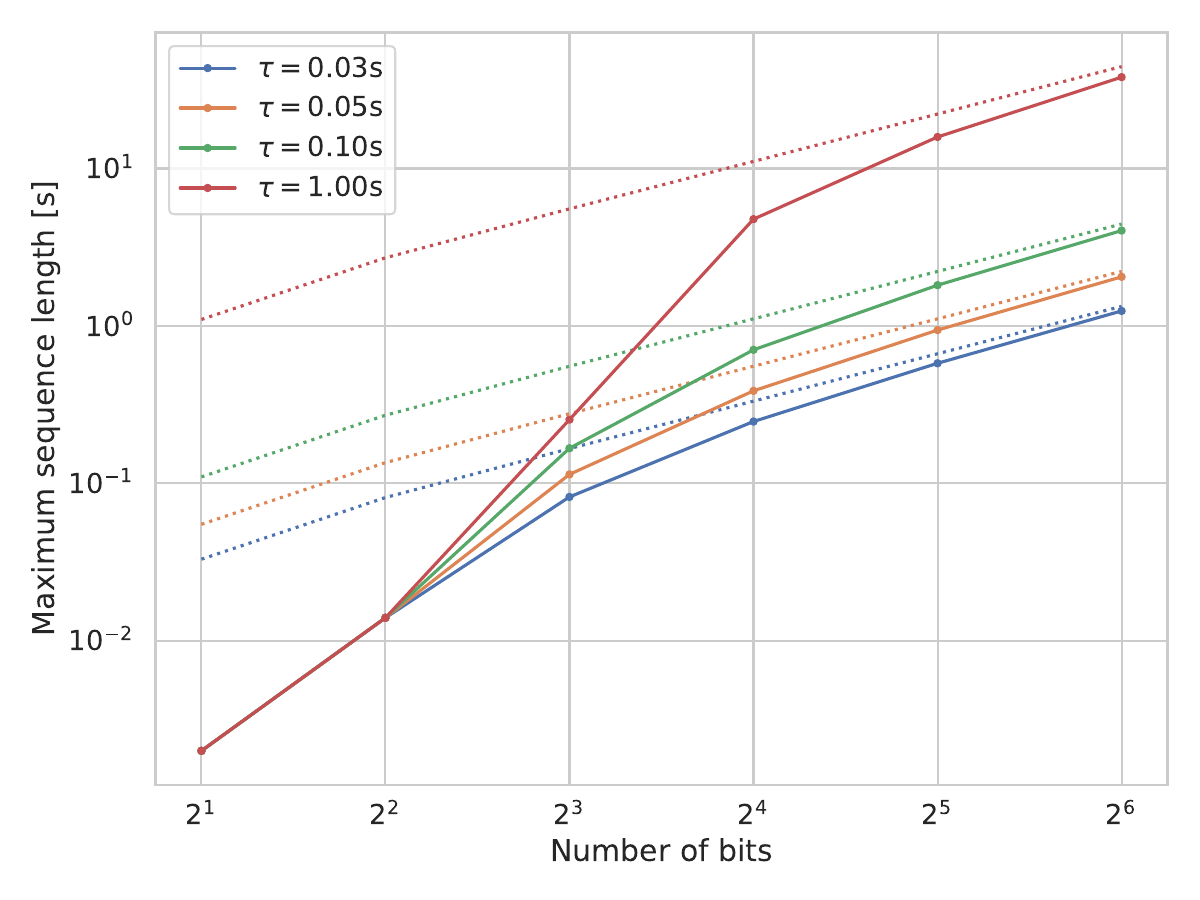}%
    \caption[Relationship between the number of bits, the leakage time constant and the maximum sequence length for a LIF neuron]{Relationship between the number of bits to encode the membrane potential, its leakage time constant $\tau$ and the maximum sequence length that can be propagated forward and backward when relying on no other mechanism, i.e. no recurrent connections. The solid line represents the empirically evaluated maximum sequence length using $\delta t=1$ms when casting back the membrane potential as an integer at each timestep. The dashed lines represent the upper-bound computed from the analytical solution. See \autoref{subsection:second_memory_lif_neuron} for details.}%
    \label{fig:digital_lif_memory_tau_vs_bits}
\end{figure}

Relying solely on path (2) limits the timescale of the task to the capacitor size of an analog neuromorphic circuit. For digital implementations, the sequence length will be limited by a relationship between $\tau$ and the number of bits available to encode the membrane potential of each neuron. Considering a standard $\delta t=1$ms, one can measure the theoretical maximum sequence length by the duration elapsed between the maximum integer value for a given number of bits ($2^{n_\text{bits}}-1$) to the minimum value of 1 during standard decay. This maximum sequence length is upper-bounded by $-\tau\log(2^{n_\text{bits}}-1)$, which is the analytical solution when not iteratively casting the membrane potential to an integer, which acts like a flooring operation. A graphical representation of this quantity is presented in \autoref{fig:digital_lif_memory_tau_vs_bits} for various $\tau$ and number of bits. We also compared the theoretical value with an empirical evaluation using standard binary representation in simulation. This resulted in lower sequence lengths, and is particularly detrimental when $n_\text{bits}<2^4$ due to the aforementioned integer casting.

\subsubsection{BPTT using the recurrent connections}

Path (1) has a denser representation in time, and therefore more reliable when dealing with tasks of longer timescales. Although, this path is subject to the same vanishing gradient problem of deep neural networks and standard RNNs. Namely, the bounds of the surrogate function $\gamma$, where $|\he'(x)|<\gamma$ for $x \in \re$, and the initialization of the recurrent weights profoundly affects how far the gradient can propagate to~\citep{Pascanu2013-va}. Furthermore, the temporally sparse objective of spiking neural networks forces the output $\vec{n}[t]$ to be mostly zeros, adding to the difficulty of backpropagating for longer timescales. Similar to how L1 and L2 regularization limits the long-term capability of the network~\citep{Pascanu2013-va}, spike-count regularization, which is usually added to enforce sparsity, will also prevent learning for longer sequences. A divergence of the network activity can therefore happen more easily during RSNN training and must therefore be dealt with.

\section{Defining a cue-accumulation task requiring long-term memory}
\begin{figure}[h]
    \centering
    \includegraphics[width=0.2\textwidth]{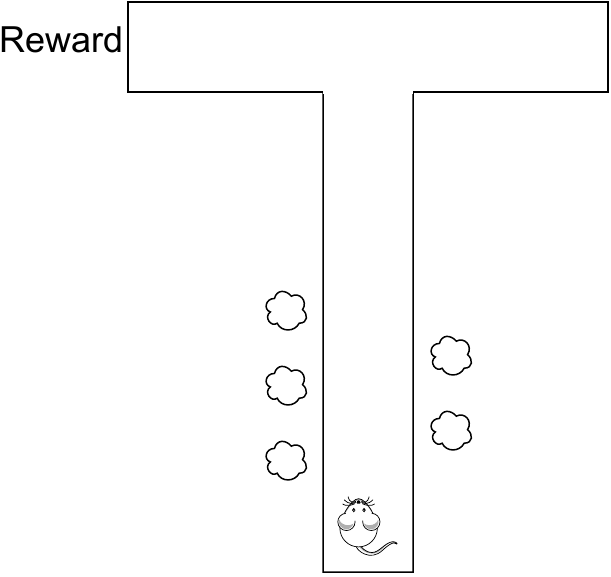}%
    \caption[Schematic representation of the cue accumulation task.]{\centering Cue accumulation task: a rodent travelling a virtual T-maze with left or right visual cues. A reward is given depending on the cue count and the direction of the rodent.}%
    \label{fig:cue_accumulation_task}
\end{figure}

To evaluate the efficiency of recurrent spiking neural networks, we consider a generated benchmark of variable length based on the neuroscience work of \citep{Morcos2016-ms,Engelhard2019-bc} and reinterpreted as an artificial benchmark in \citep{Bellec2020-ui}. In this task, a rodent travels through a T-shaped maze with visual cues displayed on the left and right walls. At the intermediate junction of the maze, the rodent has a choice of going right or left. A reward is given if the chosen direction matches the wall with the maximum number of cues. An illustration of the problem is given in~\autoref{fig:cue_accumulation_task}.

In \citep{Bellec2020-ui}'s implementation, 7 left-or-right cues are generated every 150 ms, with a 50 ms pause in between the cues. 500 to 1500 ms after the presentation of the last cue, a recall signal is provided and the network must give its decision between left or right, based on the cue count of each respective class. This recurrent network requires a minimum of $1500\ \mathrm{ms}+7\cdot 150\ \mathrm{ms}=2$ seconds of memory to solve the task. The cues are population encoded with 10 neurons per side, another 10 neurons for background noise and finally 10 neurons that provide a recall cue, to indicate the time at which a decision must be made. The recurrent spiking neural network of~\cite{Bellec2020-ui} comprises two types of neuron: leaky-integrate-and-fire (LIF) and adaptive-LIF (ALIF). The latter is a common LIF, with an adaptive threshold in which the threshold increases at every spike by a predetermined constant value, and decays exponentially to its resting state. This ALIF neuron model adds bio-plausibility since it adapts the firing response frequency of a neuron over long timescales. Another hardware-focused implementation of LSNN \cite{Frenkel2022-ij} showed that this task can be solved using a recurrent network of only LIF neurons, if the LIF’s decay constant $\tau$ is simply larger, to match the timescale of the task. Using a membrane potential leakage time constant that is as long as the adaptive threshold leakage constant of the prior work, they successfully solved this difficult temporal credit assignment task.

\subsection{The drawbacks of standard spiking recurrent neural networks}

Both \citep{Bellec2020-ui} and \cite{Frenkel2022-ij} rely on a neuron state that decays exponentially over a long period of time. In principle, this helps backpropagate the gradient over long timescales. These approaches come as an expensive cost for analog hardware implementation to store the states on capacitors. Moreover, having a variable that holds information over a long period of time hinders the fact that the network is also using recurrent connections to hold information. Indeed, while such simple tasks are often used to prove the effective memory and learning capability of a recurrent neural network, they can sometimes be solved with minimal non-linearities (e.g., the common parity problem \citep{Bengio1994-lp}). 

We create a feed-forward neural network with no recurrent connections and try to solve the task to test if the hypothesis that the increased time constants of the LIF neurons are sufficient to memorize the cues within the membrane potential of the neurons. We took inspiration from the implementation done by \citep{Frenkel2022-on}, in which we removed the recurrent connections $W_R$. This leaves us with a three-layers feed-forward network. The hidden layer uses LIF neurons with a 2-second membrane potential time constant and soft resets, such that the membrane potential gets subtracted $u_\thresh$ upon a spike:
\[
    u[t]=\exp\left(\frac{-\delta t}{\tau}\right)u[t-\delta t]+x[t]-u_\thresh n[t-\delta t]
\]

\begin{figure}[h]
    \centering
    \includegraphics[width=0.60\textwidth]{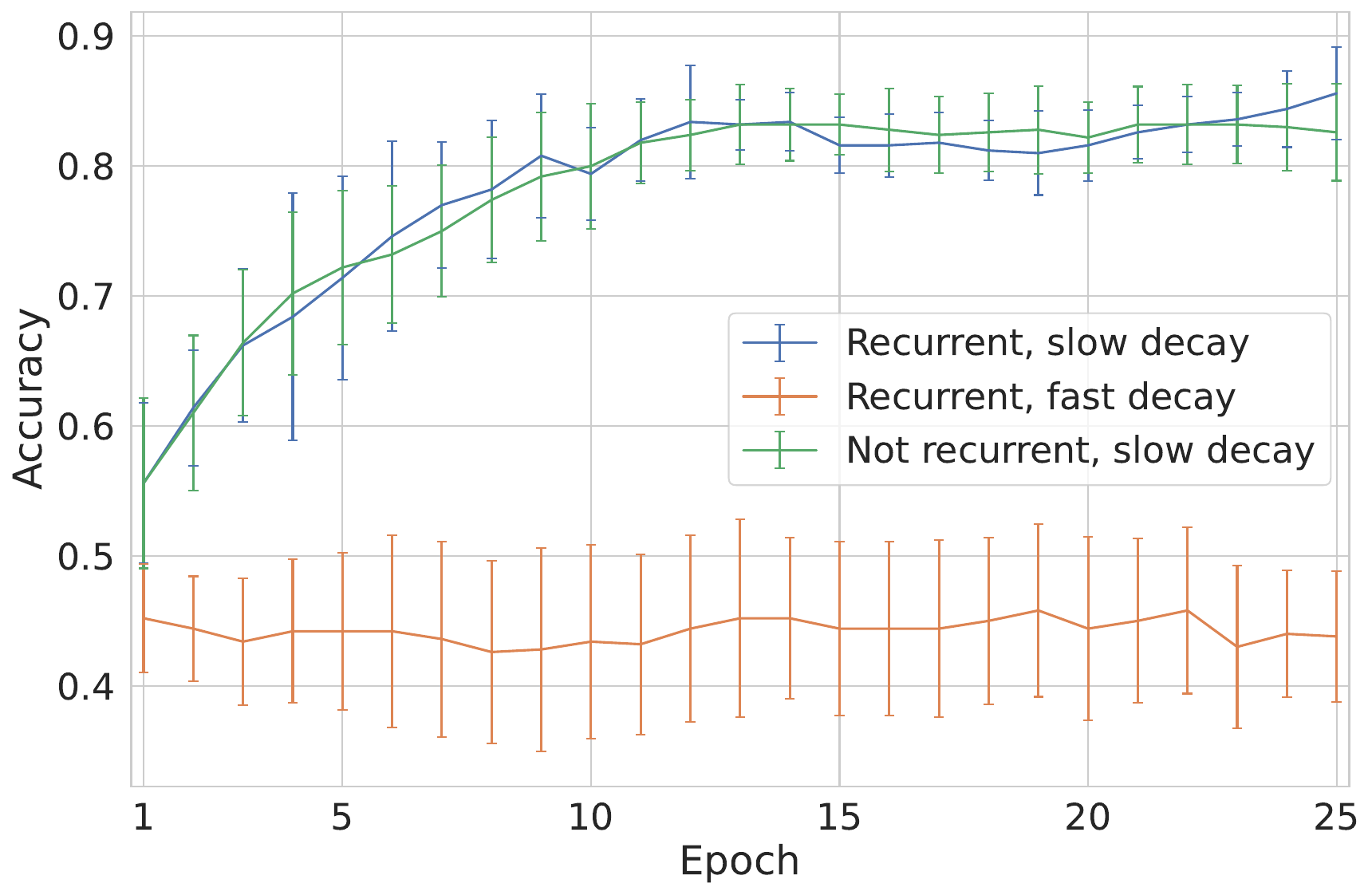}
    \caption[Accuracy of the cue accumulation task with prior methods]{Accuracy of the state-of-the-art method, implemented by \citep{Frenkel2022-on}, on the cue accumulation task in blue. We compare this network with the same network without recurrent connections in green, and with a faster leakage time constant for the neurons' membrane potentials in orange.}
    \label{fig:rsnn_vs_ffsnn}
\end{figure}

By adjusting the threshold of the neurons in the hidden layer to compensate for the diminished activity, we can achieve similar accuracy as other implementations that used recurrent connections. \autoref{fig:rsnn_vs_ffsnn} presents the accuracy of the network over 25 epochs on the cue-accumulation task. We also compare the accuracy of the network with its original recurrent implementation with slow membrane potential decay ($\tau_R=2$ seconds). All networks have 100 LIF neurons in the hidden (or recurrent) layer and are trained using surrogate BPTT, although using the e-prop learning rule yields similar results in this set-up. We also present the same network with a faster decay constant ($\tau_R=20$ ms) as comparison for a non-convergent solution, which was also observed in~\citep{Bellec2020-ui}. Similar results can be achieved with the implementation of \citep{Bellec2020-sv}, which uses 50 ALIF neurons with a long threshold adaptation time constant of 2s, and 50 LIF neurons. Once again, removing the recurrent connections in this second approach has little to no impact on the performance, which relies on the long time constant of the adaptive neurons as memory. 

Although the cue-counting task is a toy problem, it reveals flaws from the current state-of-the-art methods. Namely, increasing the time constant of the LIF neurons (or with ALIF neurons) is not only difficult to implement in analog hardware, it is possibly mitigating the adequate learning of the recurrent connections.

\section{DelayNet: a hardware-friendly spiking recurrent neural network}
In this section, we present our approach on how to solve the cue accumulation task without relying on long time constants for the leakage of a neuron state. This is done by including synaptic delays in combination with a branching-factor based regularization mechanism and a carefully designed surrogate derivative for the spike function. By doing so, we provide a network that is more hardware friendly and that can sustain longer memory. Because of the ubiquity and availability of automatic differentiation engines for training artificial neural networks, we train our networks on a clock-based numerical approximation of LIF neurons on GPU cards using PyTorch \citep{Paszke2019-eh}.

\subsection{Synaptic delays}
\begin{figure}[!h]
    \centering
    \includegraphics[width=0.7\textwidth]{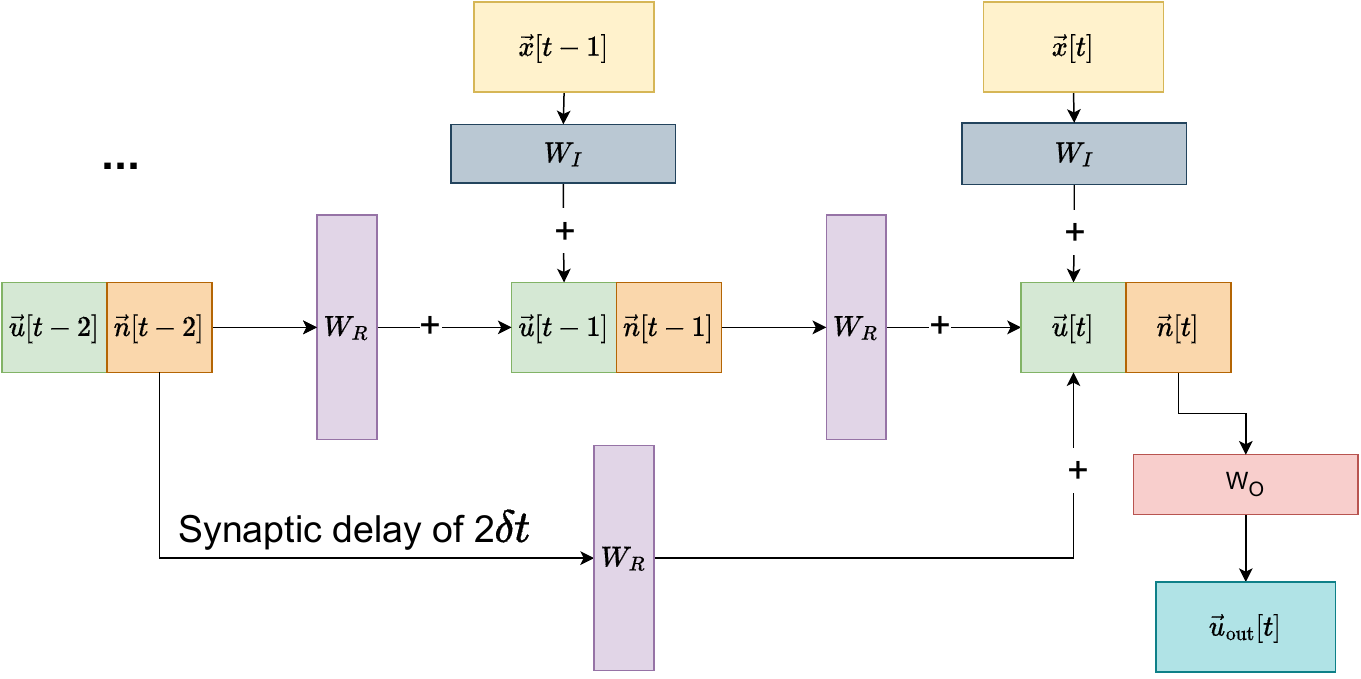}
    \caption[Computational graph of a digital RNN with delays]{Computational graph of a digital recurrent neural network with input $\vec{x}$, input weights $W_I$, recurrent weights $W_R$ and output weights $W_O$. The membrane potentials of the recurrent layer $\vec{u}$ integrate both the input, and the weighted spikes of the recurrent neurons in the same layer. When a synaptic delay is added, here a delay of 2 clock periods ($2\delta t$), the spikes are forwarded in time, creating a connection similar to residual connections in ResNets \citep{He2015-gn} and therefore helping the gradient to go further back during backpropagation through time.}
    \label{fig:skip_connection_delays}
\end{figure}

In standard digital RSNNs, there is an implicit delay of 1 timestep during spike transmission. When a neuron from the recurrent layer spikes, all the neurons connected will receive a weighted version of the spike in the next timestep. Some digital neuromorphic processors, such as Loihi~\citep{Davies2018-jo}, have the ability to increase the transmission delay by a few timesteps using a circular memory buffer. In this work, we take advantage of such mechanism to increase the memory capacity of the network. By using synaptic delays, we allow the spike transmission, and similarly the backpropagated gradient, to skip timesteps. Indeed, once a recurrent network is unravelled during backpropagation, it becomes a very deep feed forward network, with each layer representing the same recurrent layer at a different timestep. When delaying the spike transmission by a certain amount of timestep, the spike \textit{skips} a few layers and acts as if there was a direct connection between these non-consecutive layers. An illustration of that concept is presented in~\autoref{fig:skip_connection_delays}. This mechanism is conceptually similar to residual connections in ResNets \citep{He2015-gn}, but with shared weights. Residual connections played a key role in increasing the number of layers that a feed-forward network could be trained on, and a kindred concept is used here.

\begin{figure}[!h]
    \centering
    \includegraphics[width=0.5\textwidth]{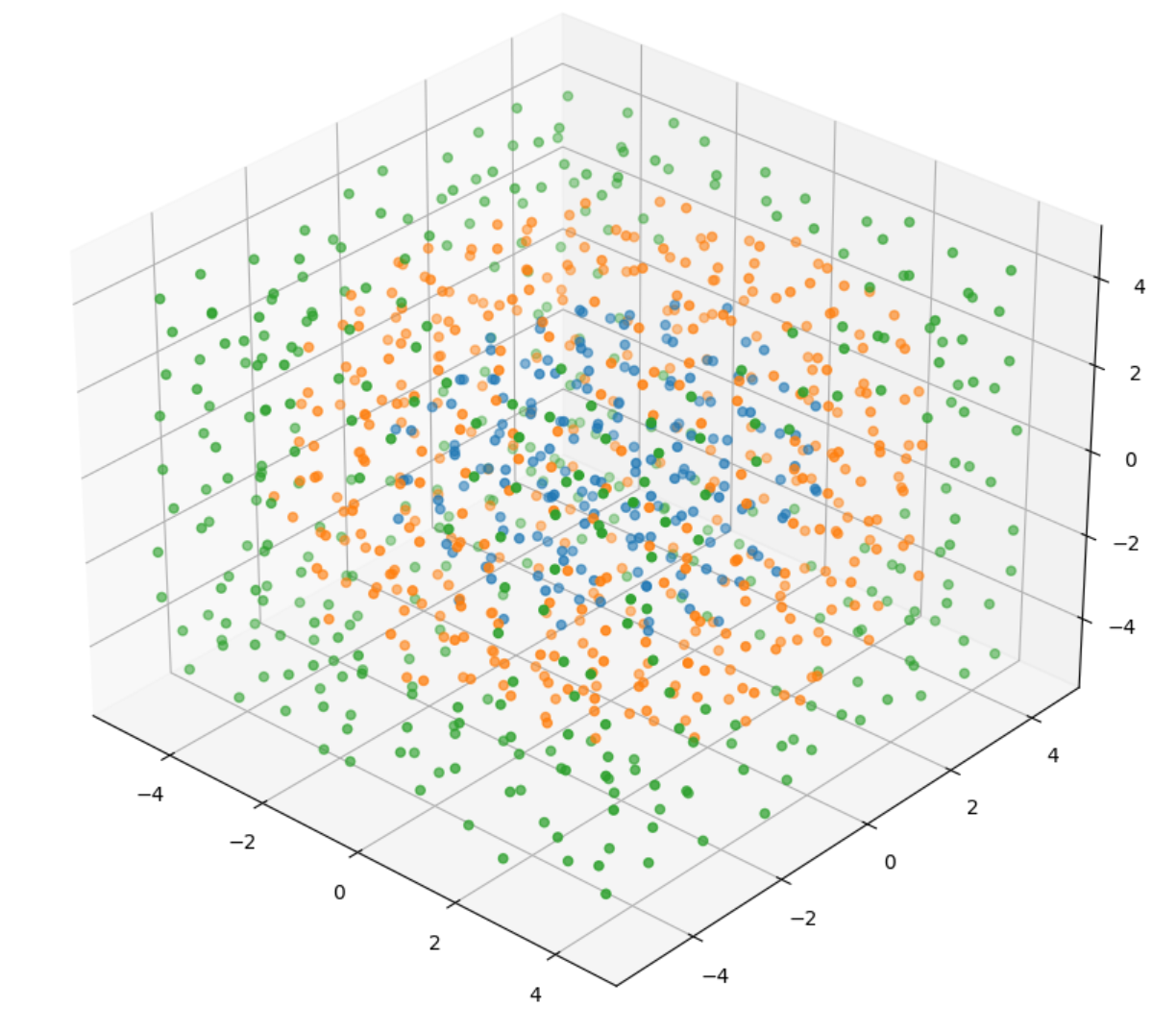}
    \caption[Three-dimensional organization of the neurons with delays]{Three-dimensional organization of one thousand neurons in a recurrent layer. Each neuron is placed in a cubic lattice and further moved by a small amount of noise. Each color represents a set of neuron with shared propagation delay for spike transmission. Neurons in green, orange and blue possess a 100, 80 and 0 ms delay respectively.}
    \label{fig:delay_position_neurons}
\end{figure}

Ideally, each pair of connections in the recurrent layer could have a propagation delay based on the distance between the pre and post neurons, and could be potentially be optimized further with e.g. brain-inspired activity-dependent myelination~\citep{Xin2020-sl}. In clock-based software simulations and analog systems, synapse-level delays are expensive. We therefore added neuron-level delays, where each neuron is associated with a fixed propagation delay, where the moment at which the neuron spikes and the moment at which the spike is propagated is delayed by a fixed constant $d$. We chose to organize the delays based on an arbitrary 3-dimensional organization of the neurons. To do so, we placed 100 neurons in a 3-dimensional space, where each neuron is positioned in a cubic lattice with positional noise uniformly sampled between 0 and 0.5. The distance between each neuron and the center reflects the delay associated with that neuron. There is a total of 3 possible delays which are 0ms for neuron within 20\% of the maximum radius, 80ms for neurons within 30\% radius and 100ms for the rest. These categories of delay were arbitrarily distributed into 17.6\%, 42.4\%  and 40.0\% of the total neuron population, respectively. A representation of the organization of the delay for each neuron is available in \autoref{fig:delay_position_neurons}. 

\subsection{Regularization with a biologically inspired branching factor}

Activity regulation is important in recurrent spiking neural network as to balance the regime of the network. Other works typically add activity regularization~\citep{Bellec2018-bd,Rossbroich2022-xn} to increase sparseness and to target or upper-bound a fixed firing rate. This firing rate is a hyperparameter that depends on the input data and potentially the number of neurons and layers. Our approach takes further inspiration from the brain, with the use of a branching factor based regularization mechanism. Recurrent neural networks are thought to operate in one of three different regimes: subcritical, critical or supercritical~\citep{Beggs2007-le}. In terms of spike transmission, each spike transmitted should average one spike produced to maintain the activity of the network and is referred to as the critical regime \citep{Haldeman2005-pa,Beggs2007-le}. Subcriticality and supercriticality occur when the average branching factor is smaller or larger than one, respectively. The brain is reported to work in a slightly subcritical regime \citep{Priesemann2014-tb}. To some extent, branching factor regularization can also replace homeostatic plasticity regularization~\citep{Rossbroich2022-xn} used to lower-bound the spiking activity, as the minimum spiking activity in the recurrent pool is dictated by the input spike train when using branching factor regularization. Astrocytes in the brain are also thought to be maintainers of this branching factor, by modulating the activity~\citep{Todd2006-it,Ivanov2021-ha}. Synaptic plasticity-enabled models have been known to improve recurrent spiking neural networks without supervised learning~\citep{Balafrej2022-ym,Ivanov2021-ha}. In this work, we compute a global branching factor based on the mean-squared error between the total spike count of each timesteps:
\[
    \xi_{bf} = \sum_{t} (\sum_i n_i[t]-n_i[t-1])^2
\]
The branching factor loss sums the activity of all recurrent neurons $n_i$ over the total duration of each sample. This loss is added to the loss of the network with a regularization parameter $\beta=0.001$:
\[
    \xi_{tot} = \xi_{net} + \beta \xi_{bf}
\] The network loss $\xi_{net}$ is the standard cross-entropy loss between the target output and the two output neurons representing the two possible paths for the rodent. Furthermore, we add standard gradient norm clipping \citep{Pascanu2013-va} with a max norm of 0.01 during BPTT to avoid exploding gradients.

\subsection{Avoiding vanishing gradient with large surrogate function}
One of the key problems of BPTT in a spiking neural network is the very temporally sparse activation function $\he$ and it's surrogate derivative $\he'$. To minimize that problem, we hypothesize that the upper bound $\gamma$ of $\he'$ such that $|\he'(x)|<\gamma$ for $x \in \re$ should be equal to one and that it's broadness\footnote{$x_\text{max}-x_\text{min}$ such that $x_\text{min}=\min_x |\he'(x)|>0$ and $x_\text{max}=\max_x |\he'(x)|>0$} must be as large as possible without being linear. This idea is similar to how the ReLU activation function has a derivative of 1 for positive values. Moreover, if the membrane potential $u$ of a neuron is ever enough to transmit a spike, i.e $u > u_{th}$, then the input of this neuron had an impact on the downstream loss function and the gradient should be propagated during backpropagation. We define $\he'$ as:
\[
    \he' = \min\{\max \{ u-u_\thresh + 1, 0 \}, 1 \}
\]
This surrogate derivative always passes the gradient backward, except if the membrane potential is highly inhibited, i.e. $u-u_\thresh + 1 < 0$.

\subsection{Weight initialization and validation}

We used the Kaiming Uniform initialization scheme~\citep{He2015-kp} on the recurrent weights $w_{r_{ij}} \in W_r$ such that the weights are sampled from a uniform distribution $-\frac{\sqrt{6}}{N}<w_{r_{ij}}<\frac{\sqrt{6}}{N}$ with $N$ the number of neurons in the recurrent pool. The diagonal (self-connections) is set to zero and forced to zero during training, i.e. $w_{r_{ii}}=0$. Fluctuation-driven initialization~\citep{Rossbroich2022-xn} is known to help enhance the memory of deep SNNs trained with surrogate gradient descent, but was unnecessary to solve the aforementioned task with DelayNet. 

\section{Results and discussion}

We tested our approach on several memory-intensive benchmarks. We first start with the aforementioned long-term memory task of \citep{Bellec2020-ui}, i.e., the cue-accumulation task. We then present novel variatiations of this task to augment the demand for memory capacity and to highlight the limitations of the current state-of-the-art approach. Finally, we examine the performance of DelayNet on the permuted-sequential MNIST task, a widely recognized benchmark across both spiking and non-spiking recurrent neural networks, to assess their memory capabilities.

\subsection{Performance of DelayNet on the cue-accumulation task}

We tested our approach on the long-term memory task of \citep{Bellec2020-ui}. In every experiment, the input matrix $W_i$, the recurrent matrix $W_r$ and the output matrix $W_o$ are trained using the MADGRAD optimizer \citep{Defazio2021-am}. Moreover, since we do not need to enforce a value for the shared membrane potential decay constant $\tau_R$ of the recurrent pool of neurons, we allow the optimizer to train $\tau_R$, along with $\tau_o$ for the output neurons, and the LIF threshold $u_\thresh$. However, all the time constants are clamped between $0.1$ ms and $50$ ms after each batch, as to not rely on the membrane potential as sufficient memory for the network. We use 125 neurons in the recurrent layer, a smaller amount than what is presented in \autoref{fig:delay_position_neurons}, but more than enough for the task. 

\begin{figure}[!h]
    \centering
    \includegraphics[width=0.9\textwidth]{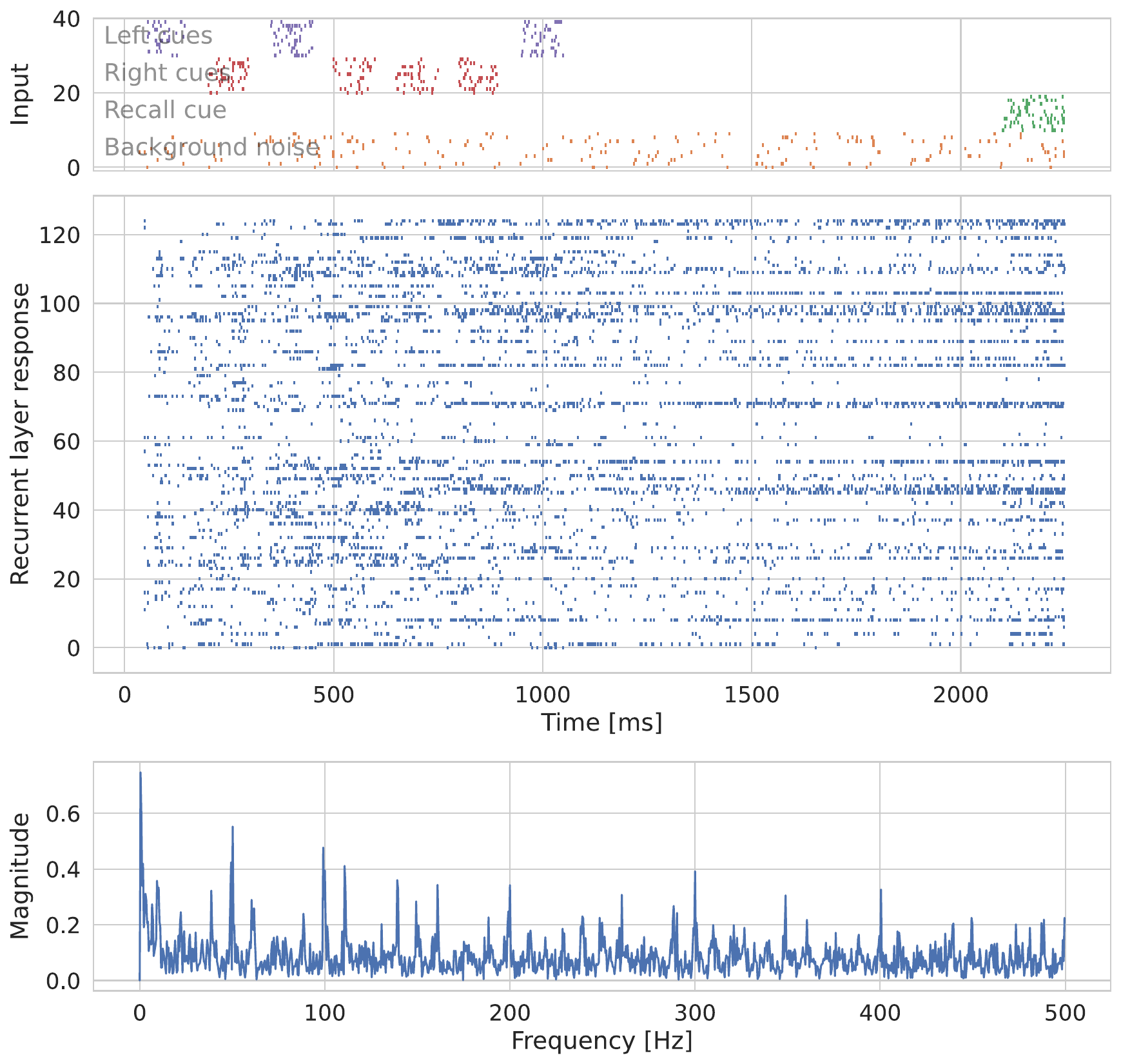}
    \caption[Network activity for one randomly selected sample of the cue benchmark.]{Network activity for one randomly selected sample of the cue benchmark. The top plot represents the input activity of the random sample, with 3 left and 4 right cues. The middle plot is the network spike activity throughout the sample. The bottom plot represents the Fourier transform of the total spike rate centered at zero. As the network remembers the input cues, oscillatory patterns start to appear at delay- and hop-dependent frequencies, with their harmonics.}
    \label{fig:spike_raster_all}
\end{figure}

We train the network over 109 epochs with 1024 randomly generated training samples per epoch. We then test on 2048 newly generated samples. A spike raster plot for one random sample is presented in~\autoref{fig:spike_raster_all}, with the spiking network activity for the recurrent layer. The network can easily solve the task with 98\% accuracy. The learned shared parameters are as follows: $\tau_R=20.61$ ms, $\tau_o=5.41$ ms, $u_\thresh$=0.72. Finally, the spectral radius of the recurrent weight matrix after learning is $1.4615$. We also present in~\autoref{fig:spike_raster_all} the Fourier transform of the spike rate summed for all recurrent neurons and centered at 0. The network showcases oscillatory patterns that are dependent on the chosen delays and their harmonics. As neurons must form loops of more than one neuron to oscillate in the network, the amplitude of the harmonics are not just artifacts. The patterns are directly representative of the information being stored in the network. Similarly, brain oscillations, or brainwaves, are thought to be a key ingredient in the brain for long-term memory and learning~\citep{Buzsaki2004-qa}. Not only is DelayNet easier to train with BPTT, we can make the comparison that it could store information similar to the brain, i.e., with oscillations.

\subsubsection{Ablation study}
We performed an ablation study to evaluate the impact of the added delays and the branching factor regularization independently. Removing either of those component prevents convergence above random accuracy. The network with branching factor regularization without delays was able to achieve 75\% accuracy on a hugely simplified task of 3 cues with $50$ ms of pause (instead of $1500$ ms).

\subsection{Variants of the cue-accumulation task}
\begin{figure}
     \centering
     \begin{subfigure}[t]{0.49\textwidth}%
         \centering%
         \includegraphics[width=\textwidth]{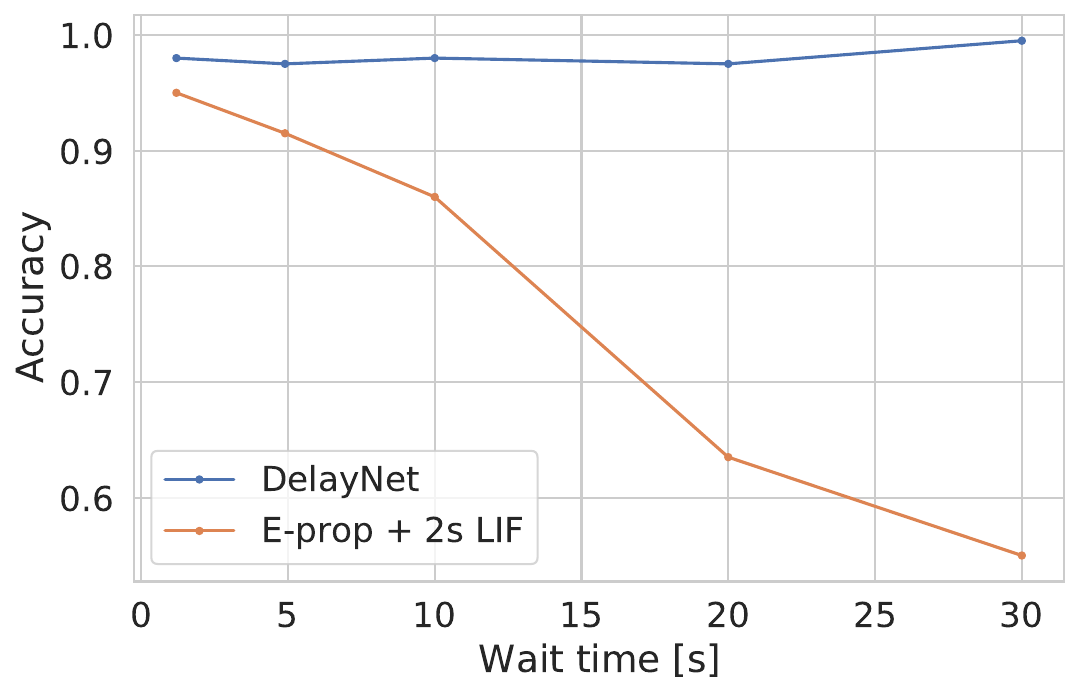}%
         \caption[Accuracy of DelayNet compared to the E-prop network]{Accuracy as a function of the wait time between the last input cue and the recall cue.}%
         \label{fig:twait_vs_acc}%
     \end{subfigure}
     \hfill
     \begin{subfigure}[t]{0.49\textwidth}%
         \centering%
         \includegraphics[width=\textwidth]{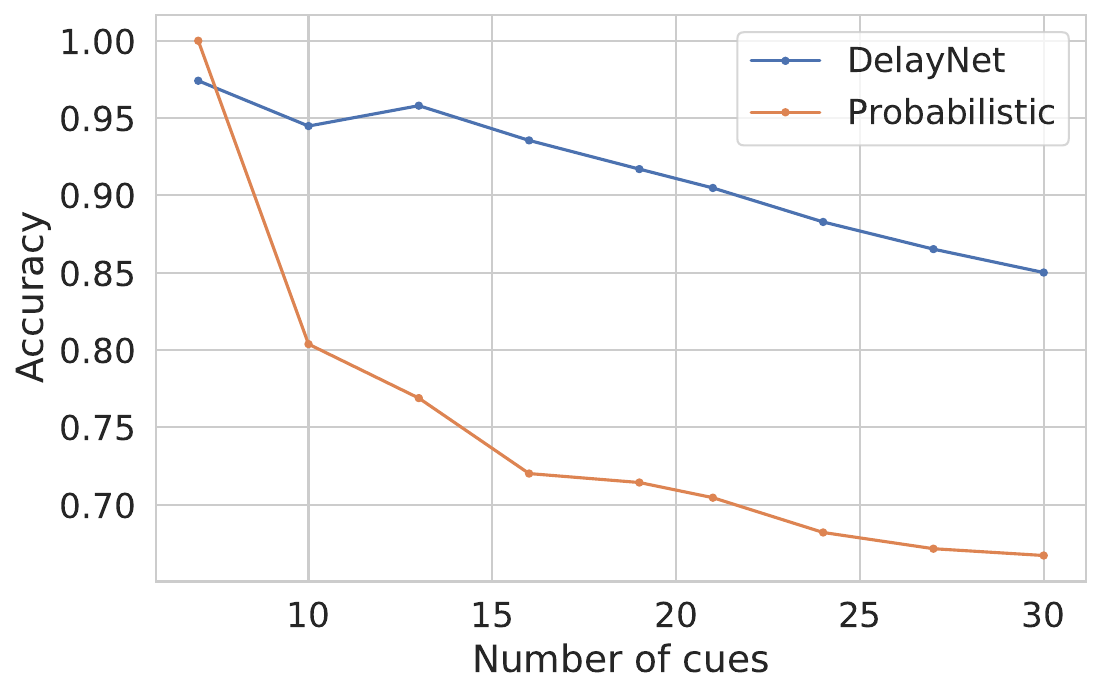}%
         \caption[Accuracy of DelayNet with an increasing number of cues]{Accuracy as a function of the number of input cues.}%
         \label{fig:nb_cues_vs_acc}%
     \end{subfigure}
     \hfill
        \caption{Accuracy of DelayNet for an increased wait period (a) and an increased number of cues (b) during inference. The wait-period showcases the limitation of an RSNN trained with E-prop and long time constants to hold the information for longer period of times. For the increasing number of cues, a probabilistic model is compared that has perfect knowledge of the last 7 cues and takes the best guess for a higher number of cue to prove that DelayNet has an increased memory.  }
        \label{fig:three graphs}
\end{figure}

The performance on the standard cue-accumulation task is not enough to showcase the network ability. We therefore created two more difficult tasks to test the limit of DelayNet. We first compare the performance of the network when increasing the waiting time after presentation of the cues. If adequate activity-based memory is achieved, the waiting time should not affect the accuracy of the network. \autoref{fig:twait_vs_acc} presents the accuracy of our network for various waiting times, compared with the RSNN implementation of~\citep{Frenkel2022-on}, with LIF neurons trained using E-prop with a long $\tau_R=2000$ms. Once again, the accuracy of the latter network quickly degrades when increasing the waiting time, as the fixed 2s time constant is not enough for such long memory period. On the other hand, our network with synaptic delays can achieve near-perfect accuracy for an arbitrary long waiting time (tested up to 30s).

We also test the ability of DelayNet to generalize over an increasing number of cues, without retraining. The performance is presented in~\autoref{fig:nb_cues_vs_acc}. To validate that the network can retain more than 7 cues, we compared the results of DelayNet with a probabilistic model. The probabilistic model consists of having a perfect knowledge of the last 7 cues, and making a decision based only on those cues. As expected, the probabilistic model can obtain above random accuracy even with more than 7 cues, but the degradation of accuracy is steeper than with the DelayNet, therefore proving that DelayNet must remember more than 7 cues. The degradation of accuracy with DelayNet is expected, as the number of neurons is fixed and the information that is contained must therefore be finite.

\subsection{Permuted sequential MNIST}

The permuted sequential mnist (psMNIST) task, as introducted in \cite{Le2015-du}, consists of taking the 784 input pixels of the MNIST handwritten digit dataset, shuffling the order of the pixels in a random but consistent order, and presenting them sequentially one at a time to a recurrent neural network for 784 timesteps. We benchmarked DelayNet on this difficult task using 216 recurrent neurons using the normalized value of the pixels as an analog input current. We split the neurons with the same ratio of delays as the cue-accumulation task, and visually presented in \autoref{fig:delay_position_neurons}. The neuron-shared delays for this task were chosen with a random hyperparameter search and have a value of 0, 25 and 86 ms. The weights and the LIF neuron threshold are then learned using the MADGRAD optimizer~\citep{Defazio2021-am}.

DelayNet showed state-of-the-art accuracy for this task when comparing to other spiking neural networks. There is still a small gap with non-spiking network due to the binary spike constraint. All the results are available in \autoref{table:psmnist_results}. 

\begin{table}[!htb]
\centering
\begin{tabularx}{0.6\linewidth}{Xrr}

\hline
Model & Spiking & Accuracy \\ \hline
SMPConv \cite{Kim2023-md} & No & \textbf{99.10} \\
M-LMU \cite{Chilkuri2021-hu} & No   &  98.49 \\
LSTM \parnote{Reported in~\cite{Chilkuri2021-hu}} & No  & 89.86 \\
Adaptive SRNN \cite{Yin2020-hl} & Yes & 91.0 \\
SRNN \cite{Bellec2018-yj} & Yes & 63.3 \\
LSNN \cite{Bellec2018-yj} & Yes & 93.3 \\
LSNN + Deep-R \cite{Bellec2018-yj} & Yes & \underline{94.7} \\
DelayNet (this work) & Yes & \underline{94.74} \\
\end{tabularx}
\parnotes
\caption{Comparison of test accuracy on \mbox{psMNIST} for both spiking and non-spiking neural networks. Best overall accuracy is in bold, best accuracy with a spiking model is underlined.}
\label{table:psmnist_results}
\end{table}

\section{Conclusion}

We showed that a simple, commonly used credit assignment task was actually insufficient to study the efficiency of spiking recurrent neural networks at solving long-term memory tasks with recurrent connections. Indeed, the state-of-the-art solutions rely on time constants that are sufficient to solve the task without recurrence, and that are not scalable on analog neuromorphic processors. We then presented DelayNet, a spiking recurrent neural network that was able to solve this task with restricted LIF decay constants. The network possesses neuron-level sub-second synaptic delays in the connections, a branching factor-based regularization mechanism, a wider surrogate function, and we provided a tool to validate the initial weights of the network. These changes allowed us to not only solve the task, but to generalize over multiple cues and increase the wait time up to 30 seconds. 

\paragraph{Limitations and future work.} Generated benchmarks like the cue-accumulation task are important to properly quantify the capabilities of spiking RNNs. They allow the quantification of the failure rate of recurrent architectures, and the training limits of such systems. In future work, we hope to provide better insights on the scalability of this approach with more realistic benchmarks. We also declare that the specific delay used (1, 80 and 100 ms) were chosen arbitrarily, and further work is necessary to provide insights on the scale of delays that can be used. We deemed them appropriate, since they are more than one scale of magnitude smaller than that length of one sample, but could possibly be reduced more.


\section*{Acknowledgements}
    We thank Damien Querlioz and Adrien Vincent for the fruitful discussions over the ideas behind this work. We acknowledge the support of the Natural Sciences and Engineering Research Council of Canada (NSERC), funding reference number 559730. Fabien Alibart acknowledges support from IONOS-ERC project \#773228. We thank the Fonds de recherche du Québec—nature et technologies (FRQNT) for the Chistera UNICO project \#287330. This research was enabled in part by support provided by Calcul Québec (calculquebec.ca) and the Digital Research Alliance of Canada (alliancecan.ca). 

{\small
\bibliographystyle{plainnat}
\bibliography{paperpile}
}


\clearpage

\appendix

\renewcommand{\figurename}{Supp. Figure} 
\renewcommand{\thefigure}{S\arabic{figure}}

\setcounter{figure}{0}

\end{document}